\documentclass{article}
\usepackage{graphicx}
\usepackage{float}
\usepackage{times}
\usepackage{dsfont}
\usepackage{algorithm}
\usepackage{caption}
\usepackage{amsmath,amssymb}
\usepackage[square,numbers]{natbib}
\usepackage{multicol}
\usepackage{siunitx}
\usepackage[bookmarks=true]{hyperref}
\usepackage{cleveref}
\usepackage{xspace}
\usepackage{xurl}
\usepackage{capt-of}
\usepackage[usenames,dvipsnames]{xcolor}
\usepackage{booktabs}  % For better-looking horizontal lines
\usepackage{threeparttable}
\usepackage{multirow}
\usepackage{times}
\usepackage{dsfont}
\usepackage{makecell}
\usepackage[normalem]{ulem}
\usepackage{colortbl}
\usepackage{xcolor}
\usepackage{svg}
\usepackage{mwe}
\usepackage{wrapfig}
\usepackage{enumitem}

\usepackage{subcaption}

\DeclareMathOperator*{\argmin}{arg\,min}

\usepackage[noend]{algpseudocode}

\definecolor{methodgreen}{RGB}{1,130,1}
\usepackage[preprint]{corl_2025}

\newcommand{\method}{{\color{methodgreen}SPI}\xspace}
\newcommand{\methodactive}{{\color{methodgreen}SPI-Active}\xspace}

\title{Sampling-Based System Identification with Active Exploration for Legged Robot Sim2Real Learning}
\author{
Nikhil Sobanbabu\textsuperscript{$\dag$1}
    \quad Guanqi He\textsuperscript{$\dag$1}
    \quad Tairan He\textsuperscript{1}
    \quad Yuxiang Yang\textsuperscript{2} 
    \quad Guanya Shi\textsuperscript{1}\\
    \textsuperscript{1}Carnegie Mellon University 
    \quad \textsuperscript{2}Google DeepMind 
    \quad \textsuperscript{$\dag$}Equal Contributions \\
    Page: \href{https://lecar-lab.github.io/spi-active_/}{\texttt{https://lecar-lab.github.io/spi-active\_/}} \\
    Code: \href{https://github.com/LeCAR-Lab/SPI-Active}{\texttt{https://github.com/LeCAR-Lab/SPI-Active}}
}
\begin{document}

\maketitle
\vspace{-5mm}
\begin{abstract}
   
Sim-to-real discrepancies hinder learning-based policies from achieving high-precision tasks in the real world. While Domain Randomization (DR) is commonly used to bridge this gap, it often relies on heuristics and can lead to overly conservative policies with degrading performance when not properly tuned. System Identification (Sys-ID) offers a targeted approach, but standard techniques rely on differentiable dynamics and/or direct torque measurement, assumptions that rarely hold for contact-rich legged systems. To this end, we present
% \comment{acronym}
    \methodactive 
    ({\color{methodgreen}S}ampling-based {\color{methodgreen}P}arameter {\color{methodgreen}I}dentification with {\color{methodgreen}Active} Exploration),
    a two-stage framework that estimates physical parameters of legged robots to minimize the sim-to-real gap. \methodactive robustly identifies key physical parameters through massive parallel sampling, minimizing state prediction errors between simulated and real-world trajectories. To further improve the informativeness of collected data, we introduce an active exploration strategy that maximizes the Fisher Information of the collected real-world trajectories via optimizing the input commands of an exploration policy. This targeted exploration leads to accurate identification and better generalization across diverse tasks. Experiments demonstrate that \methodactive enables precise sim-to-real transfer of learned policies to the real world, outperforming baselines by $42-63\%$ in various locomotion tasks.
   
\end{abstract}

\keywords{System Identification, Sim2Real, Legged Robots} 
\begin{figure*}[h]
    \centering
    \includegraphics[width=01.0\linewidth]{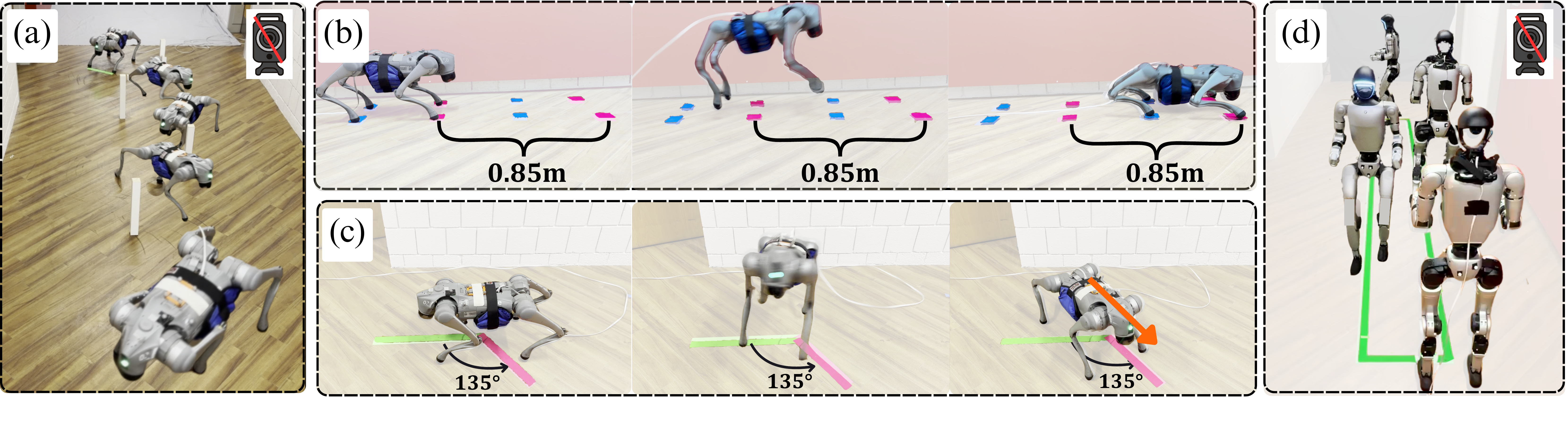}
    
    \vspace{-5mm}
   \caption{\footnotesize\methodactive enables high-fidelity Sim-to-Real transfer across diverse locomotion tasks. To highlight the precision, all tasks are open-loop tracking \emph{without global position feedback}. (a) High-Speed Weave Pole Navigation, (b) Precise Forward Jump, (c) Precise Yaw Jump, and (d) Humanoid Precise Velocity Tracking.}
   \label{fig:side_by_side}
    
\end{figure*}

% \IEEEpeerreviewmaketitle
% \vspace{-0.2cm}

% \input{sections/storyline}
\section{Introduction}
\vspace{-1mm}
Legged robots are envisioned to be used in complex environments where every stride demands a precision that leaves no room for error~\cite{lee_learning_2020,miki_learning_2022}. Reinforcement Learning (RL) has shown remarkable success in enabling agile motions on both quadruped and humanoid systems \cite{yang_cajun_2023,zhuang_humanoid_2024,he_agile_2024,hoeller_anymal_2024, zhangwococo}. However, transferring RL policies from simulation to hardware remains challenging due to the sim-to-real gap. This gap primarily stems from mismatches in physical parameters such as mass, inertia, friction, and unmodeled effects in actuator dynamics and contact interactions, where even small discrepancies can severely degrade performance in the real world. 

To bridge this gap, four broad strategies have been developed: \textbf{(1) Domain Randomization (DR)} trains robust policies by exposing them to wide parameter distributions in simulation~\cite{muratore_robot_2021,loquercio_deep_2020,peng_sim--real_2018,tobin_domain_2017}; \textbf{(2) ``White-Box'' System Identification} directly estimates physical parameters using real-world data~\cite{an_estimation_1985, mayeda_base_1990,lee_robot_2024}; \textbf{(3) ``Black-Box'' System Identification} learns full or residual dynamics model~\cite{pilco,neural_lander,hwangbo_learning_2019} from ground-truth data; and \textbf{(4) Adaptive policy learning} adapts or fine-tunes policies online using real-world feedback~\cite{kumar_rma_2021,wu2024loopsrloopingsimandreallifelong}. While practical, DR often requires heuristic tuning: excessive randomization leads to conservative policies, while insufficient randomization compromises real-world generalization. Approaches in (3) and (4) can be task-specific, prone to overfitting, and may demand substantial real-world data. In contrast, ``White-Box'' System Identification offers a principled, interpretable, and generalizable approach by estimating physically meaningful parameters, making it the focus of this work.
% 1. Existing sim2real methods can be categorized into four classes (1) DR, which XXX~cite, (2) ``white-box'' system identification, (3) ``black-box'' system ID, which XXX learning adaptive policies. 
% 2. Problems of (1,3,4). (1) Too conservative. (3,4) They are often task dependent. Compared to them, (2) gives XXX, so we focus on system id to solve sim2real in this paper.

% Third paragraph:
% 1. Challenge of system Id for legged robot: large parameter space, nondifferetiable & noncontinuous nature. 
% 2. Existing system Id not ready for this challenge
% 3. In this paper, we propose XXX
Despite its success in classic control, System Identification for legged locomotion is challenging due to severe non-linearities and intermittent contacts. Many existing methods either assume differentiable~\cite{schon_system_2011}  dynamics, rely on specialized sensing such as ground-truth torques~\cite{pfaff_scalable_2025}, or estimate only a limited subset of parameters~\cite{gautier_direct_1988, mayeda_base_1990}, limiting their applicability to general-purpose legged systems.
% These restrictions limit their applicability to general-purpose legged systems, underscoring the need for a unified, modular system identification framework deployable across platforms.
Another challenge is collecting sufficiently informative data for accurate estimation. Prior approaches often rely on hand-crafted motion scripts, simple repetitive behaviors, or isolated component tests~\cite{tan_sim--real_2018,grandia_design_2024,hwangbo_learning_2019}. While effective for subsystems, these fail to capture the coupled hybrid dynamics of natural locomotion and require task-specific tuning or extensive data collection.

% Fourth paragraph:r
% 1. Introduce the framework
% 2. Highlight key designs
% 3. Emphasize new capability
% 4. Summarize contributions

% guanqi: parallelization computation
In this work, we present \methodactive, a two-stage, parallelizable, sampling-based framework for identifying structured physical parameters of legged robots—without requiring differentiable simulators or specialized sensing. In Stage 1, we leverage heuristically designed motion priors of pre-trained RL policies to collect real-world trajectories and estimate the robot's physical parameters by minimizing state discrepancies between real and simulated rollouts.  
% pose parameter estimation as a black-box optimization problem \guanqi{a white-box optimization?}, minimizing state discrepancy between real and simulated rollouts. 
% To ensure physical plausibility and improve optimization stability, we adopt a log-Cholesky parameterization for quantities such as inertial matrices, enabling scalable parallel computation across trajectory segments. 
To enhance data efficiency and refine the initial estimates from Stage 1, Stage 2 draws on principles from optimal experiment design by maximizing the Fisher Information of the collected trajectories. Unlike prior work in  manipulators~\cite{memmel_asid_2024}, direct exploration policy training for legged robots can lead to erratic behaviors~\cite{lee_optimal_2021}. We address this by introducing a hierarchical active exploration strategy that optimizes command sequences of a pre-trained multi-behavioral RL policy—targeting informative system excitation while ensuring reliable deployment. The refined parameters significantly improve sim-to-real transfer, enabling high-precision locomotion across diverse tasks (Figure.~\ref{fig:side_by_side}) on both the Unitree Go2 quadruped and the G1 humanoid, including precise jumping, high-speed weave pole traversal, and outperforming baselines by $42-63\%$. In summary, the main contributions are: 
% in several locomotion tasks. 
% In this paper, we propose Uncertainty-Aware Sampling-Based System Identification (USS), a gradient-free method to estimate the structured physical parameters of legged robots without requiring differentiable simulators or specialized sensors. Our approach casts parameter estimation as a black-box optimization problem, where we optimize physical parameters to minimize the state discrepancy between real-world and simulated trajectories over finite time horizons. Additionally, by quantifying uncertainties in the identified parameters, we provide principled guidelines to refine DR ranges used for RL training, balancing robustness and performance without resorting to guesswork.

\begin{figure*}[t]
    \centering
    \includegraphics[width=1.0\linewidth]{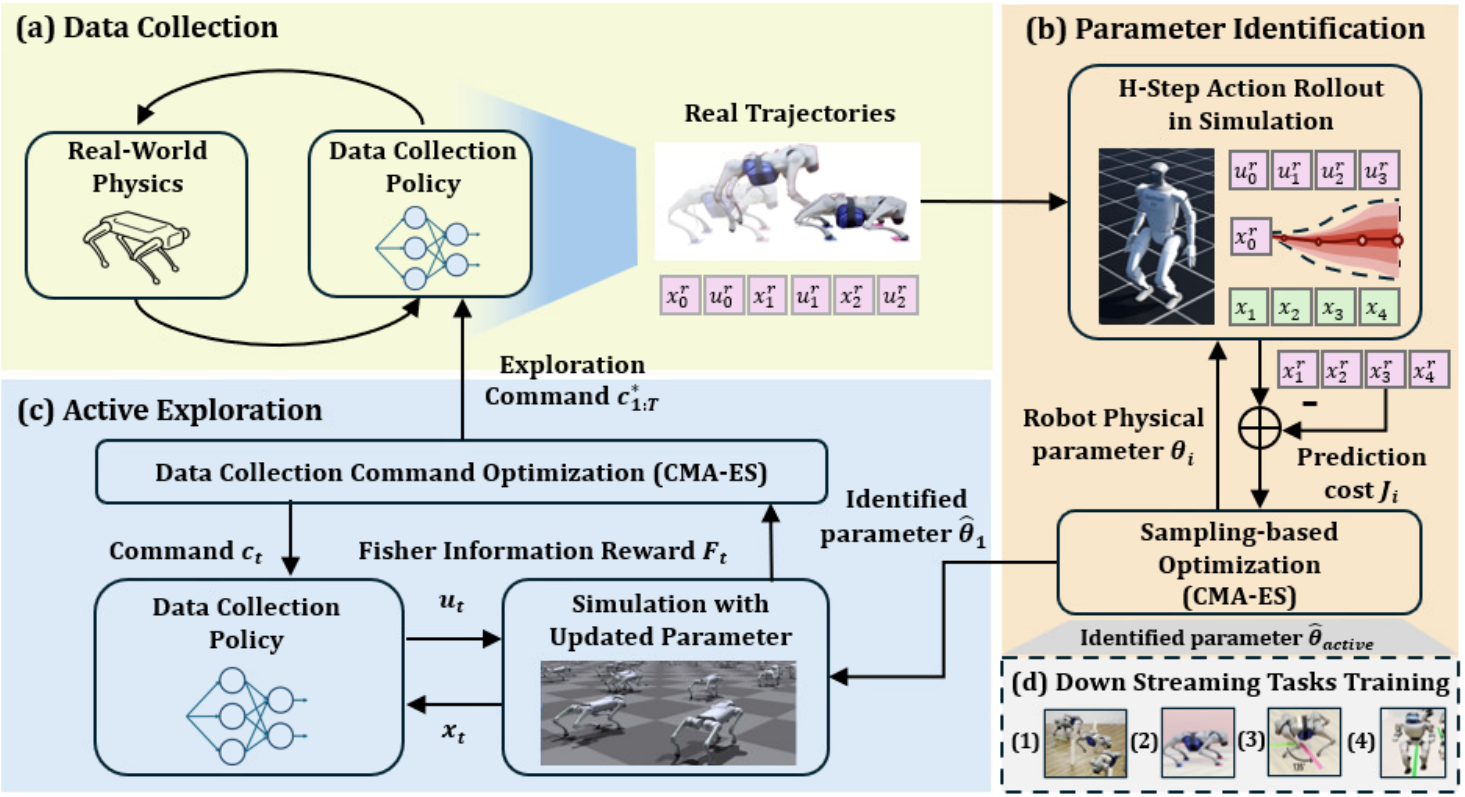}
    \caption{\footnotesize Overview of \methodactive.
\textbf{Data Collection:} Collect real-world trajectories using RL policies or motion priors.
\textbf{Parameter Identification:} Estimate physical parameters via simulation-to-real rollout matching by sampling-based optimization.
\textbf{Active Exploration:} Optimize input commands of a multi-behavioral policy to maximize Fisher Information and gather informative data.
\textbf{Downstream Task Training:} Use identified parameters to train accurate locomotion controllers.}
    \label{fig:spi-active}
    \vspace{-4mm}
\end{figure*}
% circulation, 
\vspace{-2mm}
\begin{itemize}[leftmargin=*, labelsep=0.5em]
   \item A parallelized sampling-based system identification framework for legged robots that accounts for complex contact dynamics without specialized sensor requirements.
    \item An effective active exploration strategy that optimizes the command space of a pre-trained multi-behavioral policy to induce highly informative data by maximizing Fisher Information.
    \item A comprehensive set of real-world experiments showcasing improvements in sim-to-real transfer and precise control in highly dynamic locomotion tasks for both quadrupeds and humanoids.
\end{itemize}

% \guanya{Potential hostile questions:
% \begin{itemize}
%     \item How about online adaptation?
%     \item How about learning a residual module (actuator nets, ASAP, residual dynamics)?
%     \item What is your novelty? System ID is nothing new...
%     \item Why is it related to CoRL?
% \end{itemize}
% }
\vspace{-2mm}
\section{Related Works}\label{sec:related_works}
\vspace{-1mm}
\subsection{Domain Randomization and ``Black-Box'' System Identification for Sim2Real Transfer}

A wide range of strategies have been developed to address the sim-to-real gap, including domain randomization, adaptive policy learning, and data-driven model learning.

\textbf{Domain Randomization.} Early efforts employed domain randomization (DR) to expose policies to diverse visual and physical variations during training~\cite{tobin_domain_2017}, later extending to randomized dynamics~\cite{peng_sim--real_2018, tan_sim--real_2018, sim2real_biped, mozifian2019learningdomainrandomizationdistributions, biped_gaits, he2024learning} and sensor perturbations~\cite{sadeghi_cad2rl_2016, loquercio_deep_2020}. While DR improves robustness, overly broad parameter ranges can lead to conservative policies that underperform on the true system. To address this, recent methods adapt the randomization process during training. Curriculum and adversarial DR strategies~\cite{activedr, pinto2017robust} shape the parameter distributions over time, while others refine DR bounds using real-world data~\cite{chebotar_closing_2019}. Despite these advances, DR still relies heavily on heuristics, requiring expert tuning and task-specific knowledge.

\vspace{-0.5mm}

\textbf{Learning Adaptive Policies.} Beyond DR, several approaches leverage online adaptation during deployment~\cite{kumar_adapting_2022, qi_-hand_2022, kumar_rma_2021, margolis_learning_2023, heomnih2o} or use offline data for sim-to-real transfer~\cite{bose2024offlinemultitasktransferrl} or condition the learned policy on the prediction of model parameters online~\cite{yu2017preparingunknownlearninguniversal,huang2023datt}. Some methods further adapt policies or simulation parameters during real-world rollouts to enable continual learning~\cite{wu2024loopsrloopingsimandreallifelong,smith2021leggedrobotslearningfinetuning}, but these typically require high-quality data and are not zero-shot. 
% In contrast, our method improves simulator fidelity through system identification, enabling robust generalization across tasks without relying on online fine-tuning or continual real-world adaptation.

\vspace{-0.5mm}
% \textbf{Actuator Net and Residual Model Learning.} 
\textbf{Data-driven Model Learning.}
Another class of methods learns data-driven residual networks that output corrective torques~\cite{fey2025bridgingsimtorealgapathletic} or actions \cite{he2025asapaligningsimulationrealworld} to compensate for unmodeled dynamics. While effective in specific settings, these methods risk overfitting to the training tasks or trajectories, or requiring ground-truth torques~\cite{hwangbo_learning_2019}, limiting their generalization to new or diverse tasks. 
% Our approach, by contrast, estimates physically grounded parameters, enhancing both interpretability and transferability.
\vspace{-2.0mm}
\subsection{``White-Box'' System Identification of Non-linear Dynamics}
\vspace{-1mm}

Modeling and identifying nonlinear dynamical systems remains challenging~\cite{benedetto_system_1998,natke_system_1992,schon_system_2011,o2022neural}. A foundational approach introduced least-squares estimation of inertial parameters via linearity in inverse dynamics ~\cite{an_estimation_1985}, later refined with minimal parameter sets and model selection~\cite{6DoF,industry,interative_app}. However, most methods assume structured and fixed-base models, limiting applicability to legged robots with discontinuous contacts and strong nonlinearities. Prior work mainly focuses on actuator modeling using analytical or hardware-specific approaches~\cite{tan_sim--real_2018,masuda_sim--real_2022,grandia_design_2024}, while base parameter identification often requires constrained setups or physical disassembly~\cite{tan_sim--real_2018,zhang_system_2024}. A related two-stage method~\cite{sim2real_biped} estimates actuator dynamics via latent-conditioned policies, but omits inertial parameters and lacks explicit torque decay modeling. In contrast, our framework jointly identifies inertial and actuator parameters with interpretable structure and improved sample efficiency.

\vspace{-1mm}

\subsection{Targeted Exploration for System Identification and Model Learning}
\vspace{-1mm}
% \guanya{MBRL: https://arxiv.org/abs/2306.09210}
Accurate System identification relies on collecting trajectories that sufficiently excite the dynamics of interest. Classic works on optimal experiment design~\cite{gevers_identification_2009,bombois_optimal_2011,gerencser_adaptive_2005} formalizes this using the Fisher Information Matrix (FIM) to reduce parameter uncertainty. Recent works extend this to nonlinear and hybrid systems: \cite{lee_optimal_2021} optimize excitation for mechanical systems, while \cite{sathyanarayan_exciting_2024} leverage differentiable contact simulation to identify informative contact modes. Learning-based approaches such as ASID~\cite{memmel_asid_2024} and task-oriented exploration~\cite{liang_learning_2020,wagenmaker2023optimal} actively excite dynamics via policy optimization. Others focus on scalable pipelines, including trajectory design benchmarks for inertial ID~\cite{leboutet_inertial_2021} and automated Real2Sim via robotic interaction~\cite{pfaff_scalable_2025}. Building on these foundations, we optimize command sequences of multi-behavioral policies to reliably excite informative dynamics, enabling scalable and robust parameter identification for high-dimensional legged robots.
\vspace{-1mm}
\section{\method: 
Sampling-based Parameter Identification for Legged Robot}\label{sec:sysid}
\vspace{-1mm}

% For most existing reinforcement learning pipelines, the underlying simulators—such as IsaacGym and IsaacSim—are highly parallelizable but inherently non-differentiable. To address the system identification problem in such settings, 
In this section, we introduce a zeroth-order system identification approach that leverages GPU-based parallel sampling to efficiently estimate the physical parameters of legged robots.

% \section{Model / Problem Formulation}
\textbf{Preliminary:}
% Since the rigid-body Inertia tensor variables are not independent of each other, we parametrically decompose the inertia tensor according to the work \cite{}.
% We first describe the model of the system used in our framework, followed by the reparametrization of the mass-inertia tensor that ensure physical conistency during our optimization procedure. 
Consider the parameterised dynamics of the legged system given by: $\mathbf{x}_{t+1} = f(\mathbf{x}_t, \mathbf{u}_t; \mathbf\theta)$, 
%     \begin{align}\label{dynamics_eq}
%     \mathbf{x}_{t+1} &= f(\mathbf{x}_t, \mathbf{u}_t; \mathbf\theta)
% \end{align}
where \(\mathbf{x}_t \in \mathcal{X} \subseteq \mathbb{R}^n\) is the state, \(\mathbf{u}_t \in \mathcal{U}\) is the control input, and $\mathbf\theta \in \Theta \subseteq \mathbb{R}^d$ represents the unknown model parameters to be identified using state-action trajectories collected from the real-world system. Given a dataset \(\mathcal{D}\) consisting of observed state-action sequences  $\mathcal{D} = \{(\mathbf{x}_t, \mathbf{u}_t)\}_{t=1}^{N}$, the system identification problem can be formulated as the following optimization problem:
\vspace{-1mm}
\begin{equation}\label{eq:opt-origin}
    \theta^* = \argmin_{\theta \in \Theta} \sum_{t=1}^{N} \|\mathbf{x}_{t+1} - f(\mathbf{x}_t, \mathbf{u}_t; \theta)\|^2.
\end{equation}
\vspace{-0.5mm}
% Given a dataset \(\mathcal{D}\) consisting of observed state-action sequences:
% \begin{align} \label{data-origin}
%     \mathcal{D} = \{(\mathbf{x}_t, \mathbf{u}_t)\}_{t=1}^{N},    
% \end{align}
where \(\theta^*\) is the optimal robot parameter that minimizes the prediction error between the observed states from real-world data and predicted states from the model. Specifically for legged systems, we identify a set of physical parameters \(\boldsymbol{\theta} = [\theta_{\text{in}}, \theta_{\text{mo}}]^T\), where \(\theta_{\text{in}}\) represents the mass-inertia properties, including the mass, center of mass, and inertia of rigid bodies, and \(\theta_{\text{mo}}\) denotes actuator model parameters that characterize motor dynamics. In this work, we focus on identifying the robot’s base link parameter identification. However, the approach can be readily extended to additional links with minimal modification.

\textbf{Mass-Inertia Matrix Parameterization}:
\label{mass-inertia}
% \guanya{add references}
The robot's inertial parameters $\theta_{in}$ includes: mass $m\in\mathbb{R}$, center of mass $\mathbf r\in\mathbb{R}^3$, and rotational inertia $\mathbf I\in S_3$ (the set of $3\times3$ symmetric matrices). These parameters are physically feasible if and only if the pseudo-inertia $\mathbf J(\theta_{in})$ is positive definite~\cite{sousa_inertia_2019}: $
    \mathbf J(\theta_{in}) = \begin{bmatrix}
        \mathbf\Sigma & \mathbf h \\
        \mathbf h^T & m
    \end{bmatrix} \succ 0
$, 
% \begin{equation}
%     \mathbf J(\theta_{in}) = \begin{bmatrix}
%         \mathbf\Sigma & \mathbf h \\
%         \mathbf h^T & m
%     \end{bmatrix} \succ 0
% \end{equation} 
where $\mathbf h = m \cdot \mathbf r$, $\mathbf \Sigma = \frac{1}{2}tr(\mathbf I)\mathbb{I}_{3\times 3} - \mathbf I$, and $\mathbf I_{3\times 3}$ is the identity matrix.

To satisfy the above constraints imposed on $\mathbf{J}(\mathbf\theta_{in})$, we adopt the log-Cholesky decomposition~\cite{9690029} to reparameterize the pesudo-mass matrix, thereby transforming the constrained optimization problem into an unconstrained optimization problem:
\begin{equation}
    \mathbf{J}(\theta_{in}) = \mathbf{U} \mathbf{U}^\top, \quad \mathbf{U} = e^\alpha \begin{bmatrix} 
    e_{d1} & s_{12} & s_{13} & t_1 \\
    0 & e_{d2} & s_{23} & t_2 \\
    0 & 0 & e_{d3} & t_3 \\
    0 & 0 & 0 & 1
    \end{bmatrix}
\end{equation}\label{eq:log-cholesky}
which yields a set of independently identifiable parameters:
$\boldsymbol{\phi} = [\alpha, d_1, d_2, d_3, s_{12}, s_{23}, s_{13}, t_1, t_2, t_3]^\top, \quad \boldsymbol{\phi} \in \mathbb{R}^{10}$.
% \[\boldsymbol{\phi} = [\alpha, d_1, d_2, d_3, s_{12}, s_{23}, s_{13}, t_1, t_2, t_3]^\top, \quad \boldsymbol{\phi} \in \mathbb{R}^{10}\]
The decomposition step is crucial for sampling-based optimization, as it prevents the generation of invalid mass-inertia matrices.

% \subsection{Actuator Dynamics Modeling}
\textbf{Actuator Dynamics Modeling}:\label{mtr dyn} In physics-based simulators, RL policies typically apply joint torques directly, with only torque limits enforced. However, real-world actuators exhibit significant non-linearities in high-torque regimes, leading to torque decay between commanded and actual outputs, which degrades performance in precision-critical or dynamic tasks.
To better capture these effects, we formulate an actuator dynamics model, inspired from~\cite{grandia2025design}, using a hyperbolic tangent function to map desired to actual torques in simulation: 
\vspace{-0.5mm}
\begin{equation}
\label{eq:act}
\begin{aligned}
\tau_{\text{motor}} &= \kappa \cdot \tanh\left(\frac{\tau_{\text{PD}}}{\kappa}\right), \qquad \tau_{\text{PD}}= \mathbf{K}_p (\mathbf{q}_{\text{target}} - \mathbf{q}) - \mathbf{K}_d \mathbf{\dot{q}},
\end{aligned}
\end{equation}
\vspace{-0.5mm}
where \(\mathbf{q}_{\text{target}}, \mathbf{q}, \mathbf{\dot{q}} \in \mathbb{R}^N\) denote target positions, joint states, and velocities. We assign joint-specific scaling parameters \(\kappa\) to accommodate differences in motor types and load conditions.

\textbf{Data Collection and Pre-Processing}\label{sec:data}:
% To collect system identification trajectories, we train a set of robust policies \(\{\pi_1, \dots, \pi_n\}\) with domain randomization (DR) on pre-designed tasks to excite the system dynamics parameters. Trained controllers $\pi$ is deployed to collect different trajectories forming the dataset $ \quad   \mathcal{D}_j = \{(\mathbf{x}_{t, j}, \mathbf{u}_{t, j})\}_{t=1}^{N_j}, \quad\quad \mathcal{D} = \{\mathcal{D}_j\}$.
% \begin{equation}
%     \mathcal{D}_j = \{(\mathbf{x}_{t, j}, \mathbf{u}_{t, j})\}_{t=1}^{N_j}, \quad\quad \mathcal{D} = \{\mathcal{D}_j\}
% \end{equation}
The goal of the data collection process is to induce diverse motion patterns improving parameter observability. To this end, we collect data for Stage 1 of \methodactive using heuristically designed motion-priors and diverse pre-trained RL locomotion policies (Figure.~\ref{fig:spi-active}(a)). The resulting real-word trajectories form the dataset $ \mathcal{D}_j = \{(\mathbf{x}_{t, j}, \mathbf{u}_{t, j})\}_{t=1}^{N_j}, \quad\mathcal{D} = \{\mathcal{D}_j\}$. To enhance the predictive capability of the system parameters, we extend the single-step system identification to a multi-step prediction formulation. Therefore, the dataset is segmented into various clips $\{c_k\}_{k=1}^{N_c}$ with horizon lengths $H$, 
% $    c_k = \{\mathbf{x}_{i, k}, \mathbf{u}_{i, k},\cdots,\mathbf{x}_{i+H, k}, \mathbf{u}_{i+H, k},\}$
% \begin{equation}
%     c_k = \{\mathbf{x}_{i, k}, \mathbf{u}_{i, k},\cdots,\mathbf{x}_{i+H, k}, \mathbf{u}_{i+H, k},\}
% \end{equation}
following the simulation error criterion\cite{lee_robot_2024}, where the length $H$ is sampled from uniform distribution \(\mathcal{U}(H_\text{min}, H_\text{max})\).Varying the clip length avoids bias introduced by fixed horizons and introduces an averaging effect that balances simulation error across trajectories. While this process, provides broad state-action coverage, the collected data might still lack targeted excitation of certain system parameters and designing heuristics for specific parameter excitation is often nontrivial. This limitation motivates the need for excitation trajectory design, which we address in Section.~\ref{sec:active sysid}.

\begin{algorithm}[t]
% \algsetup{linenosize=\small}
\footnotesize
\caption{\textsc{\methodactive: Two-Stage Sampling-based SysID via Active Exploration}}
\label{algo:main_algo}
\begin{algorithmic}[1]
\State \textbf{Input:} Data-collection policy $\pi(u_t \mid x_t, c_t)$, Simulator $f(x, u; \theta)$, Initial robot parameter $\theta_0$
\State \textbf{Output:} Refined parameters $\hat{\theta}_{active}$
% \vspace{-2mm}
\Statex \rule{\linewidth}{0.3pt}
\State \textbf{Stage 1: Initial Identification}
\State $\mathcal{D}_0\gets$ collect real-world data using motion priors or $\pi$ with action primitives.
\State $\hat{\theta}_1\gets$ \method($\mathcal{D}_0,\theta_0$) 
% \Comment System identification 
\State \textbf{Stage2: Active Exploration and Refinement}
\State $c_{1:T}^\star \gets \arg\min_{c_{1:T}} \ \mathrm{tr}(\mathbf{F}(\hat{\theta}_1, \pi)^{-1})$ \Comment FIM optimization using CMA-ES
\State $\mathcal{D}_1 \gets$ collect data using $\pi(u_t \mid x_t, c_t), \ c_t \sim c_{1:T}^\star$ 
\State $\hat{\theta}_{active} \gets$ \method({$\mathcal{D}_1$, $\hat{\theta}_1$})
\State \Return $\hat{\theta}_{active}$
\Statex \rule{\linewidth}{0.3pt}
\Statex \textbf{\method($\mathcal{D},\theta$)}:
\State Segment $\mathcal{D}$ into clips $\{c_k\}$, initialize CMA-ES with $\theta, \Sigma$
\Repeat
    \State Sample $\{\theta_j\}_{j=1}^B \sim \mathcal{N}(\theta, \Sigma)$ \Comment{Parallel rollouts}
    \For{each $\theta_j$}
    \State Evaluate trajectory prediction cost $J(\theta_j, \{c_k\})$ based on equation \eqref{eq:loss_fn}
        % \For{each clip $c_k$}
        %     \State $x^{\text{sim}}_{0,k} = x^{\text{real}}_{0,k}$
        %     \State $x^{\text{sim}}_{t+1,k} = f^{\text{sim}}(x^{\text{sim}}_{t,k}, u^{\text{real}}_{t,k} \mid \theta_j)$
        % \EndFor
        % \State Compute $J(\theta_j)$
    \EndFor
    \State Update CMA-ES using $J(\theta_j)$
\Until{convergence}
\State \Return $\arg\min_\theta J(\theta)$
% \vspace{-2mm}
% \Statex \rule{\linewidth}{0.3pt}

% \vspace{-1mm}
% \Statex \rule{\linewidth}{0.3pt}
\end{algorithmic}
% \vspace{-2mm}
\end{algorithm}

\textbf{Sampling-based Optimization Formulation}:
\label{sec:sysid_opti}
We formulate the identification problem as a non-linear least-squares H-Step Sequential prediction problem, with the cost function defined as:
\begin{align}\label{eq:loss_fn}
    J(\boldsymbol{\theta}, \{c_k\}) = \sum_{k=1}^{N_c} \sum_{t=0}^{H-1} \left\lVert \mathbf{x}_{t+1,k}^r - \mathbf{x}_{t+1,k} \right\rVert^2_{\mathbf{W}_x} + \lVert \theta - \theta_0 \rVert^2_{\mathbf{W}_\theta},\quad\quad
    \mathbf{x}_{t+1,k} = f(\mathbf{x}_{t,k},\mathbf{u}_{t,k}^r; \boldsymbol{\theta} )
\end{align}
where $f$ is the simulated dynamics conditioned on the parameter $\boldsymbol{\theta}$, $\mathbf{x}_{t,k}$ is the simulated state at timestep $t$ and corresponding to clip $c_k$. The initial state for each clip is aligned with the real-world trajectory segment, and subsequent states are simulated using recorded control inputs $\mathbf{u}_{t,k}^r$. Given that main-stream RL-focused simulators such as Isaacgym are non-differentiable but parallelizable, we adopt a sampling-based optimization approach inspired by the recent works~\cite{pan2024modelbased}.  The optimal parameter estimate $\hat{\boldsymbol{\theta}}$ is found by minimizing $J(\theta)$ and this optimization is performed using the CMA-ES~\cite{cmaes} framework.
At each iteration a batch of candidate parameter vectors 
$\{\boldsymbol{\theta}_j\}_{j=1}^B$ is sampled. The candidates are evaluated in parallel and $J(\boldsymbol{\theta_j})$ are  used to update CMA-ES distribution until convergence (Figure.~\ref{fig:spi-active}(b)).

\section{\methodactive: Active Exploration for Informative Data Collection}\label{sec:active sysid} 
The performance of the Sampling-based System Identification framework Section.~\ref{sec:sysid}, depends heavily on the informativeness of the collected trajectories. Although Section.~\ref{sec:sysid_opti} uses heuristic design for data collection, 
this approach cannot fully excite the system dynamics.
% this approach requires large amounts of real-world data. 
To improve data efficiency and enable more accurate parameter estimation, we introduce a principled trajectory excitation strategy that focuses on collecting a small set of highly informative trajectories.
{Cramer-Rao Bound} \cite{memmel_asid_2024} states that the covariance of any unbiased estimator $\hat{\theta}$ of the true parameters $\theta^{\star}$ is lower-bounded by the inverse of the FIM: $
\mathbb{E}_{ \mathbf{x}_{1:T}\sim p_{\theta^\star}} \left[ (\hat{\theta} - \theta^\star)(\hat{\theta} - \theta^\star)^\top \right] \succeq \mathbf{F}(\theta^\star)^{-1}$, where the FIM of the parameterized trajectory distribution $p(\mathbf{x}_{1:T} \mid \boldsymbol{\theta^\star})$ is given by: 
% \guanqi{cannot compute it, so approximate it}
\begin{equation}\label{FIM_origin}
    \mathbf{F}(\boldsymbol{\theta^\star}) = \mathbb{E}_{\mathbf{x}_{1:T} \sim p(\cdot \mid \boldsymbol{\theta^\star})} \left[ 
\left( \frac{\partial}{\partial \boldsymbol{\theta}} \log p(\mathbf{x}_{1:T} \mid \boldsymbol{\theta^\star}) \right)
\left( \frac{\partial}{\partial \boldsymbol{\theta}} \log p(\mathbf{x}_{1:T} \mid \boldsymbol{\theta^\star}) \right)^\top 
\right]
\end{equation}

% Intuitively, identifying an exploration policy $\pi_{exp}$
%   that maximizes the FIM reduces the lower bound on the estimation variance, thereby enabling more precise and reliable parameter identification. However, directly optimizing the FIM through an exploration policy poses significant challenges in the context of legged locomotion. Unlike fixed-base manipulators—where such policies can often be safely executed on hardware with minimal risk—legged robots operate under hybrid, contact-rich dynamics and are inherently more fragile. As a result, naively optimizing for FIM maximization may produce erratic behaviors that jeopardize the safety of the system, potentially leading to hardware damage or unsafe motion execution.
Intuitively, identifying an exploration policy $\pi_{exp}$
  that maximizes the FIM reduces the lower bound on the estimation variance.
  % , thereby enabling more precise and reliable parameter identification.
 However, directly optimizing the FIM through an exploration policy may produce erratic behaviors.
To this end, we propose a practical exploration strategy based on trajectory-level command optimization. Let $\pi(u_t|x_t,c_t)$ be a command-conditioned multi-behavioral policy/controller, where $u_t$ is the control action, $x_t$ is the system state and $c_t$ is the command that modulates the velocities and locomotion behaviors. Rather than learning a exploration policy from scratch, we instead optimize over the command sequences $\mathbf{c_{1:N}}$, which enables the policy to generate diverse trajectories that can excite different modes of the underlying dynamics:
\vspace{-1mm}
\begin{equation}\label{eq:active_maxim}
    \mathbf{c}^{\star}_{1:T} = \arg\min_{\mathbf{c}_{1:T}} \ \mathrm{tr}(\mathbf{F}(\theta^{\star}, \pi)^{-1})
\end{equation}
\vspace{-1mm}
% $
%     c^{\star}_{1:T} = \arg\min_{\mathbf{c}_{1:T}} \ \mathrm{tr}(\mathbf{F}(\theta^{\star}, \pi)^{-1})
% $
% \label{eq:fim_max}
% But $\mathbf{F}(\theta^{\star}, \pi)$ in \eqref{FIM_origin} is not tractable, so we approximate it using Eq:\ref{dynamics_eq}:
Considering the dynamics Eq.~\ref{eq:loss_fn}  to have a Guassian process noise,  $w_t \sim \mathcal{N}(0,\sigma^2I)$, the $\mathbf{F}(\theta^\star,\pi)$ can be approximated with~\cite{memmel_asid_2024}:
% Assuming the system dynamics as in Eq:\ref{dynamics_eq} and the trajectories are generated using a command-conditioned policy $\pi$ then the FIM reduces to: 
\vspace{-1mm}
\begin{equation}
\mathbf{F}(\theta^\star, \pi) \approx \sigma^{-2} \cdot \mathbb{E}_{p(\cdot \mid \hat{\theta}_1,\pi)} \left[ \sum_{t=1}^{T} \frac{\partial f(x_t, u_t;\hat\theta_1)}{\partial \theta} \cdot \left( \frac{\partial f(x_t, u_t;\hat\theta_1)}{\partial \theta} \right)^\top \right]
\label{eq:fim}
\end{equation}
\vspace{-1mm}
This optimization is solved using the same CMA-ES optimizer described in Section.~\ref{sec:sysid_opti}, to handle the non-differentiable dynamics and fully utilize the parallelization of the GPU-based simulator.
% however it should be noted that any relevant optimization algorithm can be used. 
Further, solving Eq.~\ref{eq:active_maxim} requires $\theta^*$ which is not available in practice, hence we substitute it with the current best estimate $\hat{\theta}_1$ obtained from the Stage 1. Additional implementation details for the FIM maximization are provided in the Appendix \ref{sec:fim_appendix}. The pseudo-code for the entire process in \methodactive is provided in the Algorithm.~\ref{algo:main_algo}. 
% \comment{have to better reference this.}
% [have a paragraph for intuition]

% \begin{algorithm}[t]
% \caption{\methodactive Two-Stage System Identification via Sampling and Active Exploration}
% \label{alg:two_stage}
% \begin{algorithmic}[1]
% \STATE \textbf{Given:} Robust policy $\pi_0$, multi-behavioral policy $\pi(u_t \mid x_t, c_t)$, simulator $f_\theta$
% \STATE \textbf{Output:} Refined parameter estimate $\hat{\theta}_2$

% \STATE \textbf{Stage 1: Initial Identification}
% \STATE $\mathcal{D}_0 \leftarrow$ collect real-world data using $\pi_0$
% \STATE $\hat{\theta}_1 \leftarrow \textsc{\method}(\mathcal{D}_0,\theta_0)$ \hfill \textit{(Algorithm~\ref{alg:optimization})}

% \STATE \textbf{Stage 2: Active Exploration and Refinement}
% \STATE $\mathbf{c}_{1:T}^\star \leftarrow \arg\min_{\mathbf{c}_{1:T}} \mathrm{tr}\left( \mathbf{F}(\hat{\theta}_1, \pi)^{-1} \right)$ using CMA-ES
% \STATE $\mathcal{D}_1 \leftarrow$ collect data using $\pi(u_t \mid x_t, c_t)$ $c_t\sim\mathbf{c}_{1:T} $
% \STATE $\hat{\theta}_2 \leftarrow \textsc{\method}( \mathcal{D}_1,\hat{\theta}_1)$ \hfill \textit{(Algorithm~\ref{alg:optimization})}

% \RETURN $\hat{\theta}_2$
% \end{algorithmic}
% \end{algorithm}

% \subsection{MLE like formulation}

% why we need hessian, we need covariance ===> DR range

% 0. Quad approximation version

% 1. have Gaussian version

% 2. improve it to exp version? 

% covarinace hessian ... derivation of the log stuff, exp distribution's MLE is equal to gamma distribution, define the UQ / covariance matrix of this 

\vspace{-1mm}
\section{Experiments}\label{experiments}
% We evaluate our framework on the Unitree-Go2 quadruped across a series of sim-to-real transfer tasks. Using Isaac Gym \cite{isaacgym}, we train reinforcement learning (RL) policies corresponding to each tasks in simulation and deploy them on the real robot to benchmark our framework against various baselines. 
% % Our analysis focuses on the role of the motor model and the impact of uncertainty quantification in system identification-based domain randomization during policy training, demonstrating its contribution to improved real-world performance. 
% We further study the importance of identifying the motor characteristics by comparing the performance of policies trained using different motor models whose parameters are identified using our framework. 
% Additionally, we conduct ablation studies on the Horizon length of datset clips $c_k$ and its effect on identification process.

\vspace{-1mm}
In this section, we present extensive experimentation results of our framework on both quadruped and humanoid systems. Through our experiments, we would like to answer the following questions:
\begin{enumerate}[leftmargin=*, labelsep=0.5em]
   %\item Does the proposed method improve the simulator’s ability to reproduce real-world dynamics more accurately?
   \item Does \method identify accurate robot models that match real-world dynamics?
   \item Do the models identified by our methods enable improved sim2real transfer of RL policies for high-precision locomotion tasks?
    % \item Does the proposed method enable the trained policies to outperform the baselines in real-world locomotion tasks? 
    \item Does the exploration strategy in \methodactive further improve the performance of \method?
    %\item Does active exploration improve prediction capability and real-world performance?
\end{enumerate}
\begin{figure}[t]
    \centering
    \includegraphics[width=1.0\textwidth]{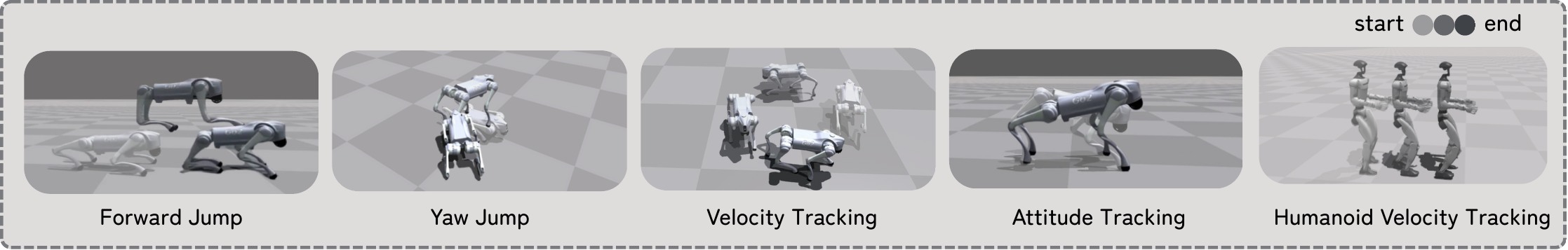}
    \caption{\footnotesize Open-Loop Locomotion Tasks: \textbf{Forward Jump}: Jump forward to a predefined distance of 0.85m. \textbf{Yaw Jump}: Jump and do Yaw Rotation to a predefined yaw angle of 135 degrees. \textbf{Velocity Tracking}: Track a sequence of Open loop 2D twist commands, \textbf{Attitude Tracking}: Track a sequence of roll and pitch commands, \textbf{Humanoid Velocity Tracking}: Track a sequence of 2D twist velocity commands for a humanoid.}
    \label{fig:tasks_def}
\end{figure}

\subsection{Tasks and Hardware Overview} 

We evaluate the framework across the platforms, Unitree Go2 
and Unitree G1. To examine the performance of sim2real transfer we consider 
four tasks namely: \textbf{Forward Jump, Yaw Jump, Velocity Tracking, Attitude Tracking} for the Unitree Go2 with an attached payload of 4.7 kg ($\sim$ one-third of its weight) and \textbf{Velocity Tracking} for Unitree G1 (Figure.\ref{fig:tasks_def}). These tasks prominently exhibit the sim-to-real gap, 
% allowing us to clearly benchmark how our framework improves performance based on predefined metrics. 
and detailed task definitions and metrics can be found in the appendix \ref{task-definition} and parameter identification results are in appendix \ref{param_analysis}. For all the tasks, RL policies were trained with Isaac Gym \cite{isaacgym} and we use Proximal Policy algorithm (PPO)~\cite{ppo} to maximize the cumulative discounted reward $\mathbb{E}\left[\sum_{t=1}^T \gamma^{t-1} r_t\right]$.

\textbf{Baselines:} For each baseline, we specify the corresponding URDF and the Domain Randomization (DR) range for the trained RL policy, if applicable:
% \vspace{-1.5mm}
% \begin{itemize}[leftmargin=*, labelsep=0.5em]
%     \item \textit{Vanilla}: This policy was trained using nominal mass-inertial parameters from the manufacturer’s URDF, with an added payload, and nominal DR ranges (Table \ref{tab:DR} column \textbf{2}).
%     \item \textit{Heavy DR}: This policy was trained with the nominal URDF+payload, but with a heavy DR range listed in table \ref{tab:DR} column \textbf{3}.
%     % \item \textit{Ours w/o UQ and motor model}: This policy was trained with URDF with idenitified base-mass inertial parameters with  with nominal DR range but without the presence of motor model.
%     % \item \textit{Ours w/o UQ}: This policy was trained with URDF with identified base-mass inertial parameters with   nominal DR ranges. However since there is no well developed intuition on how to define the ranges for the motor model parameters, we define the range to be within $(80-120)\%$ of the mean calculated. 
%     \item \textit{Gradient-based Sys-ID(\textbf{GD)}}: This policy was trained with URDF with parameters identified following \cite{pfaff_scalable_2025}\comment{Guanqi, please give more details on how this baseline was implemented, following some work will be better.}
%     \item \textit{\method}: This policy was trained with URDF with $\hat{\theta}_1$.
%     \item \textit{\methodactive}: This policy was trained with URDF with \guanqi{stage xxx result}$\hat{\theta}_{active}$.
%     % \guanqi{$\hat{\theta}_{active}$}
% \end{itemize}

\begin{itemize}[leftmargin=*, labelsep=0.5em] 
\item \textit{Vanilla}: Uses the nominal URDF with added payload and nominal DR range (Table~\ref{tab:DR}, Column \textbf{1}). 
\item \textit{Heavy DR}: Uses the same URDF as Vanilla but with a wider DR range (Table~\ref{tab:DR}, Column \textbf{2}). \item \textit{Gradient-based Sys-ID (\textbf{GD})}: URDF parameters are identified using the gradient-based optimization method \cite{9363565} with differentiable simulator (MJX \cite{todorov2012mujoco}). 
\item \textit{\method}: Uses the URDF updated with parameters $\hat{\theta}_1$ from Stage 1 and  nominal DR range. 
\item \textit{\methodactive}: Uses the URDF updated with parameters $\hat{\theta}_{active}$ from Stage 2 and nominal DR range.
\end{itemize}
% \begin{figure*}[t]
%     \centering
%     \includegraphics[width=1.0\linewidth,height=0.2\linewidth]{figs/task_def_2.png}
%     \caption{Depiction of tasks in simulation; \textbf{(i)}\textbf{Forward Jump}:Jump forward to a predefined distance, (ii)\textbf{Yaw Jump}: Jump and rotate to a predefined yaw angle, \textbf{(iii)}\textbf{Velocity Tracking}: Track a sequence of body frame planar twist commands, \textbf{(iv)} \textbf{Attitude Tracking}:Track a sequence of roll and pitch orientations, }
%     \label{fig:tasks}
    \vspace{-2mm}
% \end{figure*}
\subsection{Open-Loop Prediction using Identified Model}\label{sec:prediction_eval} 
% The performance metrics of the policies in the simulator indicate that  while the vanilla policy is able to achieve superior performance across all tasks, it severely suffers when transferred to the real world. However, our framework is able to reduce this gap and have comparable performance in sim and real. For instance, in the \textit{Omni-Directional Locomotion} task, while in the simulator the vanilla policy is able to achieve a performance metric of 0.351 by closing tracking the velocity commands and achieve a near circular trajectory, the performance severely drops in the real world leading to extreme drifts in the trajectory as shown in fig\ref{fig:side_by_side}.c. However, our framework is able to closely track the velocity commands and is able to achive the circular trajectory.This improvement is further illustrated in fig\ref{fig:sim2real}, which compares the performance of the baseline and our \method framework in both simulation and real-world execution. While the baseline policy performs well in simulation, it fails to transfer effectively to the real robot, leading to significant deviation from the desired motion. In contrast, the policy trained with \method successfully maintains consistency between simulation and reality, closely reaching the target distance of 0.85m while achieving an appropriate maximum height. This demonstrates the effectiveness of our framework in mitigating the sim-to-real gap and enabling more reliable zero-shot policy transfer.
% \vspace{-4mm}
To address \textbf{Q1}, we compare the prediction accuracy of the simulated trajectories with identified robot parameters of Unitree Go2 against real-world trajectories from a validation data. The data is collected by manually teleoperating Go2 for $60$ seconds, while it runs the RL policy following\cite{margolis2022walktheseways}.
% We compare \method and \methodactive with the system parameters corresponding to the \textit{Vanilla} and \textit{\textbf{GD}} baselines. 
We further preprocess the data, similar to Section.\ref{sec:data} segmenting the data in various clips of average horizon length of 1.5sec. 
We evaluate the prediction accuracy by comparing the mean tracking error of the global root position $J_{rpos}$, per-joint angle $J_{pja}$ and the global root velocity $J_{rvel}$. We report the normalized Quantitative results in Table.~\ref{tab:performance}(a) demonstrates that \method and \methodactive  consistently outperform the baselines, achieving lower 
$J_{rpos}$ and $J_{pja}$, indicating more accurate open-loop prediction and closer alignment with real-world dynamics. 
In contrast, the gradient-based method \textit{\textbf{GD}} exhibits larger prediction errors, possibly due to non-differentiable contact dynamics and the sim-to-sim gap between MJX and Isaac Gym.  
% \comment{what other insights do we have from this exp, anything about GD basedline?}

\vspace{-1mm}

\subsection{Sim2Real Performance}
\vspace{-1mm}
\begin{wrapfigure}{r}{0.65\textwidth}
    \vspace{-4mm}
    \centering
    \includegraphics[width=\linewidth]{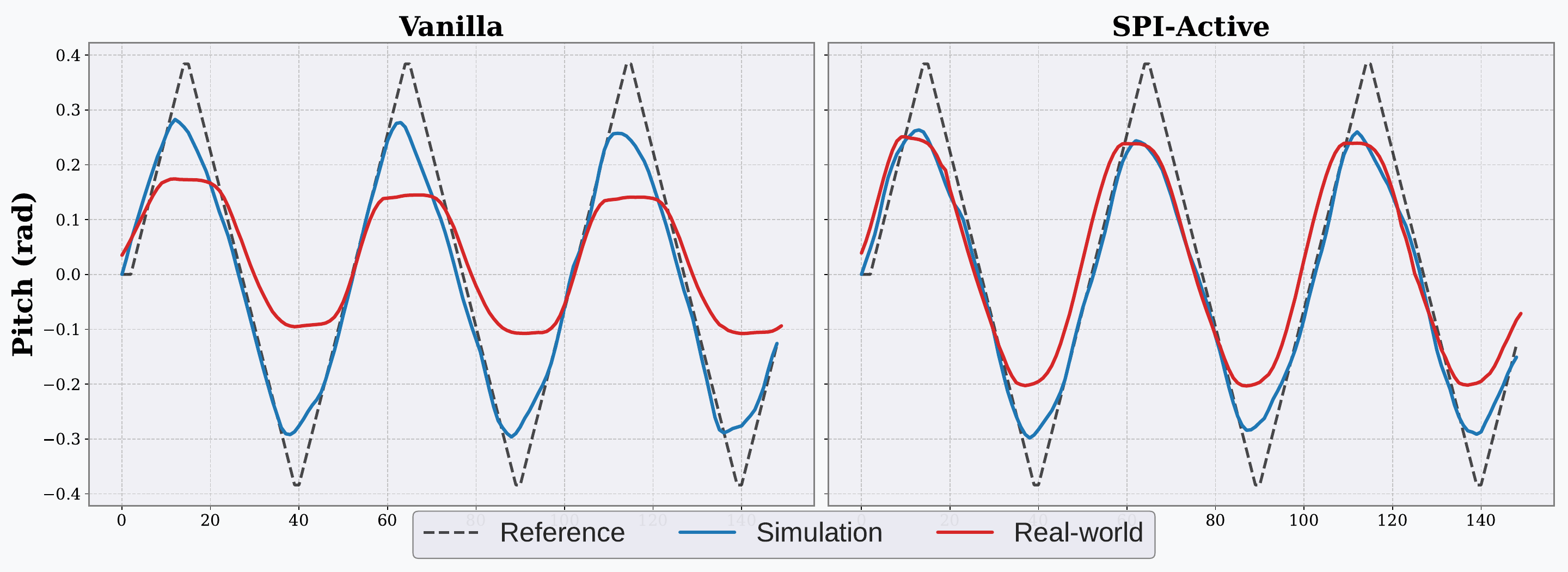}
    \vspace{-3mm}
    \caption{\footnotesize  Comparison between SPI-Active and Vanilla policies in both simulation and real-world execution. SPI-Active yields a closer match to simulation, suggesting improved sim-to-real consistency.}
    \label{fig:attitude_tracking}
    \vspace{-2mm}
\end{wrapfigure}

\begin{figure*}[t]
    \centering    \includegraphics[width=0.9\linewidth]{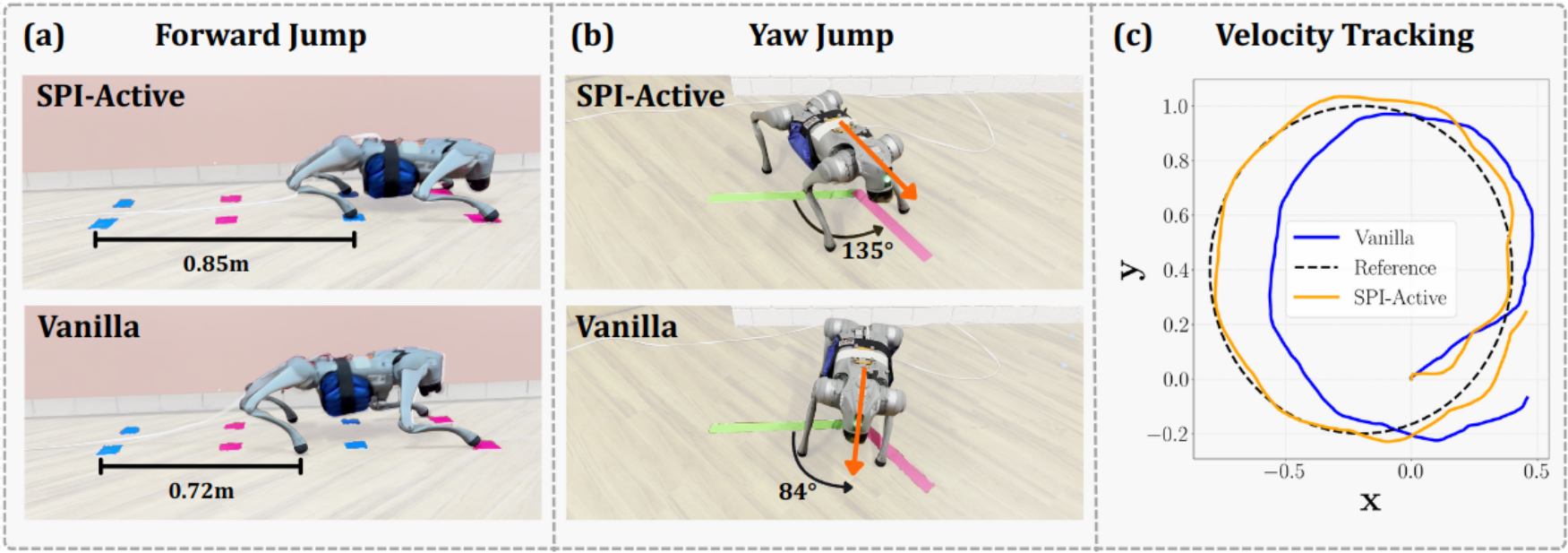}
    \caption{Task Performance comparison of \methodactive vs Vanilla in (i) Forward Jump, (ii) Yaw Jump, (iii) and Velocity Tracking}
    \label{fig:tasks}
    \vspace{-5mm}
\end{figure*}

To address \textbf{Q2}, we evaluate RL policies fine-tuned with system parameters identified by each method.  Policies are trained for tasks described in Appendix \ref{task-definition} and deployed directly on hardware without further tuning. 
% As reported in Table~\ref{tab:performance}(b), both \method and \methodactive consistently outperform the baselines across all tasks. In \textit{Velocity Tracking}, \textit{Forward Jump} and \textit{Yaw Jump}, \method achieves an improvement of 19.6\%, 39.9\% and 35.9\% respectively over the \textit{Vanilla} baseline, demonstrating improved sim-to-real transfer. While \textit{Heavy DR} outperforms \method in the \textit{Attitude Tracking} task - likely due to lesser excitation of attitude related system parameters in the Stage 1 dataset, \methodactive surpasses all baselines across every task. In particular, it achieves a 63.6\% gain in the \textit{Yaw Jump} task, underscoring the benefit of targeted excitation via active exploration. Additionally, both \method and \methodactive improve performance in the \textit{Humanoid Velocity Tracking} task, demonstrating the framework’s generalization across robot morphologies and tasks.
As reported in Table~\ref{tab:performance}(b), both \method and \methodactive consistently outperform the baselines across all tasks. In \textit{Velocity Tracking}, \textit{Forward Jump} and \textit{Yaw Jump}, \method achieves an improvement of 19.6\%, 39.9\% and 35.9\% respectively over the \textit{Vanilla} baseline, highligting improved sim-to-real transfer (Figure.~\ref{fig:tasks}). While \method under performs  in the \textit{Attitude Tracking} task - likely due to insufficient excitation of attitude-related system parameters, \methodactive surpasses all baselines in \textit{Attitude Tracking} (Figure. \ref{fig:attitude_tracking}) and across every task emphasizing the effectiveness of targeted excitation. Additionally, \method improves performance in the \textit{Humanoid Velocity Tracking} task, demonstrating the framework’s generalization across robot morphologies and tasks.

\begin{table}[t]
\centering

\caption{(a) Comparison of Prediction Accuracy and (b)Task Performance Across Methods }
\vspace{-1mm}
\begin{subtable}[t]
{0.45\textwidth}\centering
\caption{ Normalized Aggregate Prediction Error}
\resizebox{0.74\textwidth}{!}{%
\begin{tabular}{lccc}
\toprule
\textbf{Method}  &\(J_{rpos}\downarrow\)&\(J_{pja}\downarrow\) & \(J_{rvel}\downarrow\)\\
\midrule
Vanilla            & 1.00&1.00&1.00  \\
GD  & 0.98  & 1.14&1.05  \\
\method  & 0.87&0.89& 0.95\\
\methodactive &\textbf{0.67} &\textbf{0.72}&\textbf{0.91} \\
\bottomrule
\end{tabular}
}
\end{subtable}
\hfill
\begin{subtable}[t]{0.54\textwidth}
\centering
\caption{ Normalized Performance Across Tasks (\(\downarrow\))}
\resizebox{\textwidth}{!}{%
\begingroup
\setlength{\tabcolsep}{4pt}
\renewcommand{\arraystretch}{0.9}
\begin{tabular}{lccccc}
\toprule
\textbf{Method} 
& \(J_\text{fj}\) 
& \(J_\text{yj}\) 
& \(J_\text{vt}\) 
& \(J_\text{at}\)
& \(J_\text{hvt}\) \\
\midrule
Vanilla & 1.00{\tiny\(\pm\text{0.000}\)} 
& 1.00 {\tiny\(\pm\text{0.000}\)}
& 1.00 {\tiny\(\pm\text{0.000}\)}
& 1.00 {\tiny\(\pm\text{0.000}\)}
& 1.00 {\tiny\(\pm\text{0.066}\)}\\
Heavy DR    & 1.75{\tiny\(\pm\text{0.033}\)} 
& 1.40 {\tiny\(\pm\text{0.021}\)}
& 1.06 {\tiny\(\pm\text{0.022}\)}
& 0.85{\tiny\(\pm\text{0.040}\)}
& 1.10 {\tiny\(\pm\text{0.064}\)}\\
\method & 0.60{\tiny\(\pm\text{0.014}\)} 
& 0.64 {\tiny\(\pm\text{0.066}\)}
& 0.80 {\tiny\(\pm\text{0.044}\)}
& 0.96{\tiny\(\pm\text{0.042}\)}
& 0.87{\tiny\(\pm\text{0.064}\)} \\
\methodactive & \textbf{0.48}{\tiny\(\pm\text{0.012}\)} 
& \textbf{0.37} {\tiny\(\pm\text{0.038}\)}
& \textbf{0.58} {\tiny\(\pm\text{0.014}\)}
& \textbf{0.73}{\tiny\(\pm\text{0.052}\)}
& \textbf{-}  \\ 
\bottomrule
\end{tabular}
\endgroup
}

\end{subtable}
\vspace{-3mm}
\label{tab:performance}
\end{table}

\begin{figure*}[t]
    \centering    \includegraphics[width=1.0\linewidth]{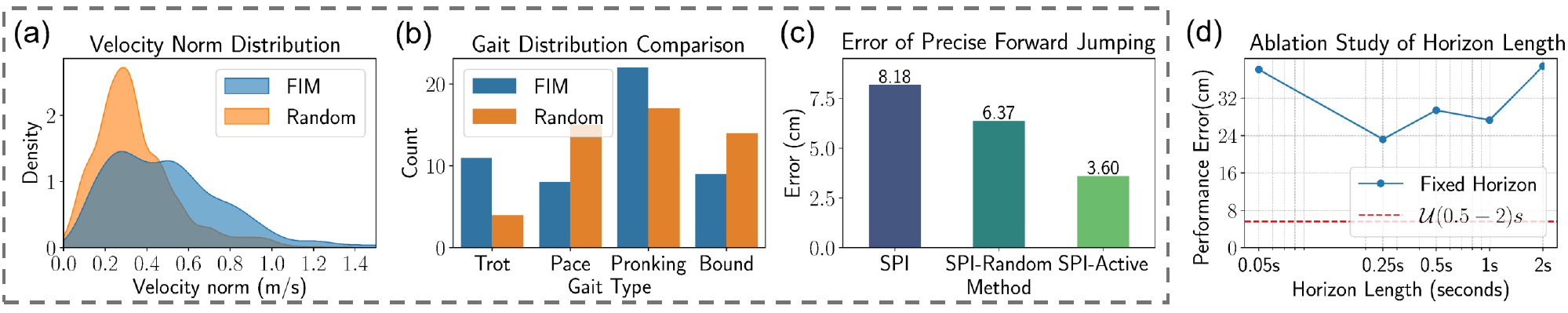}
    \caption{Effect of Active Exploration: (a) Velocity Magnitude Distribution (b) Gait Pattern Distribution and (c) Forward Jumping error comparison. Horizon Length Ablation (d)} 
    \label{fig:active_exp}
    \vspace{-4mm}
\end{figure*}
\subsection{Performance Improvement with Active Exploration}
\vspace{-3mm}
To investigate the effect of active exploration (\textbf{Q3}),
we evaluate the impact of active exploration on real-world task performance by comparing three variants on the \textit{Forward Jump} task: (1) \method, (2) \method+random where second-stage data is collected using randomly sampled input commands, and (3) \methodactive. As shown in Fig.~\ref{fig:active_exp}(c), \methodactive yields a jumping distance error (3.6cm), lower compared to the other variants, demonstrating the effectiveness of targeted trajectory excitation for parameter identification. Further, Fig.~\ref{fig:active_exp}(a) and (b) demonstrate that FIM-based command optimization induces trajectories with higher velocity norm distributions and a higher occurrence of high-torque gaits such as pronking. This results in richer excitation of the system dynamics, which is critical for accurate parameter identification.

\vspace{-2mm}
\subsection{Ablation Studies}
\vspace{-2mm}
\textbf{Role of Actuator Modeling in Policy Transfer:}
\label{sec:motor-model}
Here we discuss the performance impact of using different motor models.
We evaluate four torque models: (1) \textbf{Vanilla} (ideal torques), (2) \textbf{Linear Gain}, (3) \textbf{Unified Tanh}  with unified $\kappa$, and (4) \textbf{Ours} (motor-specific $\kappa$), where we choose unique $\kappa_i$ for hip, calf and thigh motors of Unitree Go2. Real-world experiments were conducted on the \textit{Forward Jump} and \textit{Yaw Jump} tasks. We report the normalized task metrics with respect to the Vanilla baseline in Table \ref{tab:motor_model_ablation}.
\begin{wraptable}{r}{0.55\textwidth}
\scriptsize
\vspace{-3mm}
\centering
\caption{\scriptsize Comparison of Motor Models on Normalized Task Metrics}
\setlength{\tabcolsep}{2pt}
\begin{tabular}{lccc}
\toprule
\textbf{Method} & \textbf{Definition} ($\tau_{\text{motor}}\sim$) & $J_{fj} \downarrow$ & $J_{yj} \downarrow$ \\
\midrule
Vanilla & $\tau_{PD}$ 
& 1.00{\tiny\(\pm\text{0.00}\)} 
& 1.00{\tiny\(\pm\text{0.00}\)} \\
Linear Gain & $\kappa \cdot \tau_{PD}$ 
& 1.30 {\tiny\(\pm\text{0.023}\)}
& 1.54 {\tiny\(\pm\text{0.011}\)}\\
Unified Tanh & $\kappa \cdot \tanh(\tau_{PD} / \kappa)$ 
& 0.74 {\tiny\(\pm\text{0.009}\)}
& 1.17 {\tiny\(\pm\text{0.019}\)} \\
Ours & $\mathbf{\kappa_{i}} \cdot \tanh(\tau_{PD} / \kappa_i)$
& \textbf{0.55} {\tiny\(\pm\text{0.015}\)}
& \textbf{0.76}{\tiny\(\pm\text{0.042}\)} \\
\bottomrule
\end{tabular}
\label{tab:motor_model_ablation}
\vspace{-2mm}
\end{wraptable}
It can be observed that, our model achieves the lowest error in both tasks. In the \textit{Forward Jump} task, our method and Unified Tanh outperform the vanilla baseline by 45\% and 26\% respectively, highlighting the benefit of modeling joint-specific nonlinearities for high-torque maneuvers. However, in the \textit{Yaw Jump} task, the Unified Tanh underperforms the Vanilla baseline, which is likely due to the lower torque demands and the effect of angular inertia, where unified scaling fails to capture joint-specific behavior.

% \begin{wrapfigure}{R}{0.4\textwidth}
%     \centering
%     \includegraphics[width=0.4\textwidth]{figs/horizon_length_ablation.png}
%     \caption{Performance comparison for different Horizon Lengths(s) of the dataclips. Our strategy of Uniformly sampled horizon lengths shows significant performance improvements compared to the estimations made with fixed horizons}
%     \label{fig:horizon_ablation}
%     % \vspace{-2mm}
% \end{wrapfigure}

\vspace{-2mm}
\textbf{Influence of Horizon Length on Policy Performance:} We also investigate the effect of horizon length \(H\) when segmenting trajectory clips \(c_k\) during system identification, using the \textit{Forward Jump} task. We compare fixed-length horizons (0.05s to 2s) against uniformly sampled horizons within this range. As shown in Figure~\ref{fig:active_exp}(d), uniformly sampled horizons lead to better estimation and downstream performance.
Results show that smaller \(H\) fails to capture long-term temporal dependencies needed for accurate estimation, while larger \(H\) suffers from instability of the open-loop action rollout. Uniformly sampling \(H\) during Stage 1 yields a better trade-off, enabling the identification process to benefit from both short and long-horizon dynamics. 

\section{Conclusion}\label{sec:conclusion}
\vspace{-3mm}
We presented a two-stage, sampling-based system identification framework for legged robots that combines robust physical parameter estimation with an active trajectory excitation strategy to enable precise and scalable sim-to-real transfer. Our method does not rely on differentiable simulators or ground-truth torques, making it broadly applicable across different robotic platforms. By leveraging heuristic RL policies in the first stage and optimizing command sequences for Fisher Information in the second, our approach generates informative trajectories that excite key inertial and actuator parameters ensuring reliable hardware execution. Experimental results on the Unitree Go2 and G1 demonstrate significant improvements in tracking accuracy and task performance compared to domain randomization and baseline identification approaches. These results highlight the importance of accurate model identification and targeted data collection in bridging the sim-to-real gap for legged locomotion.

\section{Limitations}\label{sec:limitations}
While our framework demonstrates strong performance on quadrupeds, its application to humanoids is currently limited to Stage 1 identification. Extending active trajectory excitation to humanoid systems is a promising direction, but presents challenges due to their high dimensionality and safety-critical dynamics. Additionally, our method operates offline; enabling online or adaptive identification could improve real-time performance in dynamic settings. The approach also assumes access to a multi-behavioral policy for generating diverse motions, which may not generalize to novel morphologies. Lastly, while CMA-ES enables parallelizable optimization, it can be computationally demanding in high-dimensional spaces, motivating future work on more sample-efficient or uncertainty-aware alternatives.

%% Use plainnat to work nicely with natbib. 
\section{Acknowledments}
We would like to thank Yuanhang Zhang for his support in training locomotion policies for the Unitree G1.

\bibliography{references}

\clearpage
\appendix
\section{Appendix}
% \guanqi{fix the section layout}
\subsection{Tasks Definition} \label{task-definition}
% \begin{figure}[h]
%     \centering
%     \includegraphics[width=1.0\textwidth]{figs/task_def_2.png}
%     \caption{Open-Loop Locomotion Tasks: \textbf{Forward Jump}: Jump forward to a predefined distance of 0.85m. \textbf{Yaw Jump}: Jump and do Yaw Rotation to a predefined yaw angle of 135 degrees. \textbf{Velocity Tracking}: Track a sequence of Open loop 2D twist commands, \textbf{Attitude Tracking}: Track a sequence of roll and pitch commands, \textbf{Humanoid Velocity Tracking}: Track a sequence of 2D twist velocity commands for a humanoid.}
%     \label{fig:enter-label}
% \end{figure}
\subsubsection{\textbf{Forward Jump}}  
The forward jump task requires the robot to jump forward to a predefined horizontal distance \(x\). For our experiments, we set \(x = 0.85~\text{m}\). Additionally, during the training process, the policy was incentivized to maximize the vertical jump height, reaching a target height of \(z = 0.35~\text{m}\) above the initial height of the robot.  
To quantitatively evaluate the performance of the forward jump task, we define the performance error metric as:  
\begin{align*}  
J_{\text{fj}} = \lvert x_f - x_i - 0.85 \rvert + \lvert y_f - y_i \rvert + \lvert \max z - z_i - 0.35 \rvert,  
\end{align*}  
where \((x_i, y_i, z_i)\) and \((x_f, y_f, \max z)\) represent the robot's initial and final positions and the maximum height achieved during the jump, respectively. 
This metric captures the robot's ability to achieve the desired forward displacement while maintaining lateral stability (\(y_f - y_i\)) and achieving the target vertical jump height (\(0.35~\text{m}\)).  

\subsubsection{\textbf{Yaw-Jump}}
The yaw jump task requires the robot to perform an in-place jump while achieving a specified yaw rotation. For our experiments, the target yaw angle was set to \(\frac{3\pi}{4}\) radians. The performance error metric is defined as:  
\begin{align*}  
J_{\text{yj}} = \lvert \phi_f \rvert + \lvert \theta_f \rvert + \lvert \psi_f - \psi_i - \frac{3\pi}{4} \rvert + \lvert x_f - x_i \rvert + \lvert y_f - y_i \rvert,  
\end{align*}  
where \(\phi_f\), \(\theta_f\), and \(\psi_f\) represent the final roll, pitch, and yaw angles in the robot's body frame, respectively, \(\psi_i\) is the initial yaw angle, and \((x_i, y_i)\) and \((x_f, y_f)\) denote the robot's initial and final horizontal positions. This metric accounts for the robot's ability to maintain stability in roll and pitch, achieve the desired yaw rotation of \(3\pi/4\) radians, and minimize horizontal drift during the jump.  
 
\subsubsection{\textbf{Velocity Tracking}}
 In this task, the policy was trained to track velocity commands consisting of forward velocity (\(v_x\)), lateral velocity (\(v_y\)), and angular velocity (\(w_z\)). To evaluate the policy, we designed a circular trajectory tracking task, where the forward velocity command is varied linearly: it increases from \(0.4~\mathrm{m/s}\) to \(0.8~\mathrm{m/s}\) and then decreases back to \(0.4~\mathrm{m/s}\). The angular velocity command was adjusted accordingly to achieve a circular trajectory of radius \(0.6~\mathrm{m}\). 
The performance metric for this task is defined as:  
\begin{equation}  
J_{\text{vt}} = \sqrt{(v_{x,\text{ref}} - v_x)^2 + (v_{y,\text{ref}} - v_y)^2} + 0.5 \cdot \lvert w_{z,\text{ref}} - w_z \rvert,  
\end{equation}  
where \(v_{x,\text{ref}}\), \(v_{y,\text{ref}}\), and \(w_{z,\text{ref}}\) are the reference forward, lateral, and angular velocities, and \(v_x\), \(v_y\), and \(w_z\) are the actual tracked velocities. We follow the same definitions for the Humanoid Velocity Tracking task.

\subsubsection{\textbf{Attitude Tracking}} 
In the attitude tracking task, the policy was trained to achieve commanded roll or pitch angles, all defined with respect to the body frame. During evaluation, the task involved tracking a periodic-ramp pitch reference signal with a fixed amplitude and frequency, followed by a periodic-ramp roll reference signal.  
The performance metric for this task is defined as:  
\begin{equation}  
J_{\text{rp}} = \sum_{i=1}^{N} \left\|  
\begin{bmatrix}  
\phi^{ref}_{i} - \phi_{i} \\  
\psi^{ref}_{i} - \psi_{i}  
\end{bmatrix}  
\right\|_2,  
\end{equation}  
where \( \phi^{ref}_{i} \) and \( \psi^{ref}_{i} \) are the sinusoidal reference signals for roll and pitch, respectively, and \( \phi_{i} \) and \( \psi_{i} \) are the actual tracked roll and pitch angles over the evaluation period.

\subsection{Implementation Details of \method}\label{spi_details}

The system identification objective is defined by the dynamics model in Eq.~\eqref{eq:opt-origin}. The state vector $\mathbf{x}_t$ represents the full floating-base and joint state of the robot, defined as \[\mathbf{x}_t = [\mathbf{p}_t, \mathbf{q}_t, \mathbf{v}_t, \boldsymbol{\omega}_t, \mathbf{q}_{\text{jnt},t}, \dot{\mathbf{q}}_{\text{jnt},t}]\] where $\mathbf{p}_t \in \mathbb{R}^3$ is the base position, $\mathbf{q}_t \in \mathbb{R}^4$ is the base orientation represented as a unit quaternion, $\mathbf{v}_t \in \mathbb{R}^3$ is the linear velocity of the base, $\boldsymbol{\omega}_t \in \mathbb{R}^3$ is the angular velocity, $\mathbf{q}_{\text{jnt},t}$ denotes joint positions, and $\dot{\mathbf{q}}_{\text{jnt},t}$ denotes joint velocities.

The system identification cost function Eq.\eqref{eq:loss_fn} in \method consists of three components: \textbf{Base Prediction Cost}, which promotes global pose alignment by penalizing errors in the simulated floating-base state relative to motion-capture trajectories; \textbf{Joint Prediction Cost}, which enforces local dynamic consistency by minimizing discrepancies in joint position, velocity, and torque using proprioceptive data; and \textbf{Parameter Regularization}, which constrains deviations from nominal URDF values for physical and motor parameters, including mass, center of mass, inertia, and actuator gains.

% The system identification cost function \eqref{eq:loss_fn} in \method comprises three categories of terms: \textbf{Base Prediction Cost}, \textbf{Joint Prediction Cost}, and \textbf{Parameter Regularization}.

% The cost function comprises three components: a \textbf{Base Prediction Cost} that enforces global pose alignment by penalizing errors in the simulated floating-base state against motion-capture trajectories; a \textbf{Joint Prediction Cost} that ensures local dynamic consistency by minimizing discrepancies in joint position, velocity, and torque using proprioceptive sensor data; and a \textbf{Parameter Regularization} term that constrains deviations from nominal URDF values for physical and motor parameters, including mass, center of mass, inertia, and actuator gains.

% \textbf{Base Prediction Cost} penalizes discrepancies between the simulated floating-base state and the reference trajectory recorded by the motion-capture system, including errors in position, velocity, orientation, and angular velocity, to promote global pose alignment.

% \textbf{Joint Prediction Cost} penalizes mismatches between simulated and measured joint states, including position, velocity, and torque. Reference values are obtained from the robot’s proprioceptive sensors, ensuring local dynamic consistency and capturing contact dynamics.

% \textbf{Parameter Regularization} constrains deviations from nominal values specified in the URDF, applied to mass, center of mass, diagonal inertia, and motor model parameters. Default motor gains are selected based on actuator torque limits.

The actuator model in Eq.\eqref{eq:act} employs a hyperbolic-tangent form to approximate torque saturation. It preserves undisturbed torque output under small commands while smoothly saturating at the limits, ensuring both physical realism and optimization stability.

The full set of cost terms and their corresponding coefficients is detailed in Table~\ref{tab:cost}. Coefficients are first normalized to yield unit cost on a reference dataset using default parameters, followed by global scaling: velocity-related terms are weighted by $0.5$, torque-related terms by $0.2$, and regularization terms by $0.1$.

The parameter sampling ranges for CMA-ES initialization are detailed in Table~\ref{tab:sampling_range}, where each parameter is drawn from a uniform distribution centered at its nominal value specified in the URDF. Sampling-based optimization is performed using Optuna~\cite{akiba2019optunanextgenerationhyperparameteroptimization} with the default CMA-ES optimizer with Gaussian sampler, running for 5 iterations.

\begin{table}[h]
\centering
\small
\caption{\method Cost Function Terms and Coefficients}\label{tab:cost}
\begin{tabular}{llc}
\hline
\textbf{Name} & \textbf{Expression} & \textbf{Coefficient} \\
\hline
\multicolumn{3}{l}{\color{black}{Base prediction cost}} \\
\hline
Position prediction error & $\|p - p_r\|^2$ & 4.0\\
Velocity prediction error & $\|v - v_r\|^2$ & 2.0\\
Quaternion prediction error & $1.0 - \langle q, q_r \rangle^2$ & 2.0\\
Angular velocity prediction error & $\|\omega - \omega_r\|^2$ & 0.5\\
\hline
\multicolumn{3}{l}{\color{black}{Joint prediction cost}} \\
\hline
Joint position prediction error & $\|q_{jnt} - q_{jnt,r}\|^2$ & 3.0\\
Joint velocity prediction error & $\|\dot q_{jnt} - \dot q_{jnt,r}\|^2$ & 0.1\\
Joint torque prediction error & $\|\tau - \tau_{r}\|^2$ & 0.01\\
\hline
\multicolumn{3}{l}{\color{black}{Parameter regularization}} \\
\hline
Mass & $\|m - m_0\|^2$ & 0.01\\
Center of mass  & $\|\textbf r - \textbf r_0\|^2$ & 10.0\\
Inertia & $\|\textbf I - \textbf I_0\|^2$ & 1.0 \\
Tanh motor gain & $\|\kappa_{\tanh} - \kappa_{\tanh,0}\|^2$ & 0.01\\
Linear motor gain & $\|\kappa_{s} - \kappa_{s,0}\|^2$ & 0.1\\
\hline
\end{tabular}
\end{table}

% \subsubsection{Cost Function Component}

% cost terms and cost coefficient (table)\\
% Identification cost function \eqref{eq:loss_fn} of \method consists of the following terms:

% \subsubsection{Parameter Sampling Range}
\begin{table}[h]
\centering
\small
\caption{Sampling Ranges of Parameters for Different Robots}\label{tab:sampling_range}
\begin{tabular}{lcc}
\hline
\textbf{Name} & \textbf{Min} & \textbf{Max} \\
\hline
\multicolumn{3}{l}{\color{black}{Go2 (base link)}} \\
\hline
Mass $m$ & $3.0$ & $15.0$ \\
Inertia diagonal elements $\text{diag}(\textbf I)$ & $(0.005, 0.005, 0.005)$ & $(1.0, 1.0, 1.0)$ \\
Center of mass $\textbf r$ & $(-0.1, -0.1, -0.1)$ & $(0.1, 0.1, 0.1)$  \\
Tanh motor gain $\kappa_{\tanh}$ & $(10.0, 10.0, 10.0)$ & $(40.0, 40.0, 40.0)$ \\
Linear motor gain $\kappa_{s}$ & $0.5$ & $1.5$ \\
\hline
\multicolumn{3}{l}{\color{black}{G1 (pelvis link)}} \\
\hline
Mass $m$ & $1.0$ & $10.0$ \\
Inertia diagonal elements $\text{diag}(\textbf I)$ & $(0.005, 0.005, 0.005)$ & $(1.0, 1.0, 1.0)$ \\
Center of mass $\textbf r$ & $(-0.2, -0.2, -0.2)$ & $(0.2, 0.2, 0.2)$  \\
\hline
\end{tabular}
\end{table}

% Hyper parameter of CMA-ES
% \subsubsection{Hyper parameter of CMA-ES}

\subsection{Implementation Details of \methodactive}\label{sec:fim_appendix} \label{spi_active_details}
The active exploration in Stage-2 of \methodactive requires us to optimize the input command sequences of a multi-behavioral policy(Section.\ref{sec:active sysid}). To this end, we follow the training pipeline of \cite{margolis2022walktheseways} and pre-train an RL policy whose input commands $c_t$ is a 14 dimensional vector, given by: 

\begin{equation}
    c_t = [v_x,v_y,w_z,h,f,b_1,b_2,b_3,b_4,h_f,\phi,\psi,s_w,s_l]^T
\end{equation}
where $v_x,v_y,w_z$ are the 2D velocity twist commands, and the policy is trained to track these commands, $h,f,h_f$ refers the body height, stepping frequency and the foot swing height respectively. $b_1,b_2,b_3,b_4$ are the gait behavioral commands, that modify the quadrupedal gait and out of which $b_4$ determines the duration of the gait which is kept fixed even during the training phase. $\phi,\psi$ are the body roll and pitch commands, and $s_w,s_l$ are the commanded stance width and length. 

In order to solve the optimization problem in Equation.\ref{eq:active_maxim}, we need to  approximate the value of $\mathrm{tr}(\mathbf{F}(\theta^{\star}, \pi)^{-1})$ for a given trajectory. Hence, we consider our dynamics equation with a gaussian noise as given by: 

\begin{equation}
   \mathbf{x}_{t+1} = f(\mathbf{x}_t, \mathbf{u}_t;\mathbf{\theta}) + w_t
\end{equation}
where $w_t \sim \mathcal{N}(0,\sigma^2I)$, then the FIM reduces to:
\begin{equation}
    \mathbf{F}(\theta^\star, \pi) =  \sigma^{-2} \cdot \mathbb{E}_{p(\cdot \mid \theta^\star,\pi)} \left[ \sum_{t=1}^{T} \frac{\partial f(x_t, u_t;\theta^\star)}{\partial \theta} \cdot \left( \frac{\partial f(x_t, u_t;\theta^\star)}{\partial \theta} \right)^\top \right]
\end{equation}
Now, given that $\theta^\star$ is not available to us, we instead use a surrogate with $\hat{\theta}_1$ and given that our simulators need not be differentiable, we use finite difference approximation to calculate the gradient. Further, we add a termination penalty to prevent highly aggressive inputs commands that can lead to fall of the robot.  

% \begin{itemize}
%     \item talk about freezing some commands
%     \item optimizing gait behaviors as a whole 
%     \item spline interpolations to have high frequency and make it smoother
%     \item horizons within T. 
% \end{itemize}

Further, If we want to collect a trajectory of length $\sim 40s \implies T=2000$, to make this optimization problem more tractable, we constrain the input space by optimizing only a selected subset of commands: $[v_x, v_y, \omega_z, b_1, b_2, \phi, \psi]$ while keeping others fixed. These were chosen for their ability to sufficiently excite the physical parameters of interest during system identification. It is important to note that this subset is not fixed and can be adapted based on the specific parameters being targeted in different scenarios. Rather than optimizing these commands at every individual timestep, we adopt a more compact representation by reparameterizing the command trajectories using a 10-degree Bézier curve. This reduces the number of optimization variables, as we only sample and optimize the corresponding control points of the Bézier curve, excluding $b_1$ and $b_2$. The entire command sequence is divided into segments of fixed time horizons, each of length $H = 4$ seconds. For each horizon, the Bézier-defined command profile is resampled to generate the control sequence.For the gait-modulating commands $b_1$ and $b_2$, we select from four discrete combinations: $(0.5, 0.5)$, $(0.5, 0.0)$, $(0.0, 0.5)$, and $(0.0, 0.0)$, which correspond to pace, trot, bound, and pronk gaits respectively. These gait parameters are held fixed within each horizon to preserve consistent behavioral structure during execution.

% \subsection{Mass-Inertia Matrix Parameterization}\label{sec:logchol}

% % \subsection{Mass-Inertia Matrix Parameterization}
% % \textbf{Mass-Inertia Matrix Parameterization}:
% \label{mass-inertia}
% % \guanya{add references}
% The robot's inertial parameters $\theta_{in}$ includes: mass $m\in\mathbb{R}$, center of mass $\mathbf r\in\mathbb{R}^3$, and rotational inertia $\mathbf I\in S_3$ (the set of $3\times3$ symmetric matrices). These parameters are physcially feasible if and only if the pseudo-inertia $\mathbf J(\theta_{in})$ is positive definite~\cite{sousa_inertia_2019}: $
%     \mathbf J(\theta_{in}) = \begin{bmatrix}
%         \mathbf\Sigma & \mathbf h \\
%         \mathbf h^T & m
%     \end{bmatrix} \succ 0
% $, 
% % \begin{equation}
% %     \mathbf J(\theta_{in}) = \begin{bmatrix}
% %         \mathbf\Sigma & \mathbf h \\
% %         \mathbf h^T & m
% %     \end{bmatrix} \succ 0
% % \end{equation} 
% where $\mathbf h = m \cdot \mathbf r$, $\mathbf \Sigma = \frac{1}{2}tr(\mathbf I)\mathbb{I}_{3\times 3} - \mathbf I$, and $\mathbf I_{3\times 3}$ is the identity matrix. 
% To ensure that these physical constraints are satisfied, we project the mass-inertia matrix $\mathbf{J}$ into a log-Cholesky form \cite{9690029} that is feasible with respect to the positive-definite constraints.
% % To satisfy the above constraints imposed on $\mathbf{J}(\mathbf\theta_{in})$, we project the mass-inertial matrix to a LMI-feasible log-Cholesky form to make sure the matrix follows the physical constraints. 

\subsection{Implementation details for Training Downstream Tasks}
The training pipeline for all the downstream tasks uses  the code framework inspired from \cite{HumanoidVerse}. We first mention the commanalities and then report the task specific rewards, observations and details. For each task, we formulate it as a goal-conditioned Reinforcement learning (RL) task, where the policy $\pi$ is trained to achieve a task and motivated to reduce the performance metrics in Appendix.\ref{task-definition}.Each policy is conditioned by observations $o_t$ and it outputs action $a_t\in \mathbb{R}^{12}$ for the quadruped and $a_t \in \mathbb{R}^{15}$ for the humanoid, where the policy provides actions only to the lower body and the upper body joints are fixed.  These actions correspond to the target joint positions and is passed to a PD controller that actuates the robot's degrees of freedom. The policy uses PPO \cite{ppo} to maximize cumulative discounted reward. Further, we follow an asymmetric actor-critic training \cite{he2025asapaligningsimulationrealworld} to train the actor with only easily available observations from proprioception and other time based observations, while the critic has access to privileged information like base linear velocity which is usually difficult to estimate with on-board sensors. 

\subsubsection{Observations}
All the policies use the robot's proprioception $s^p_t$ and some task specific observations. The \textit{Forward Jump} and \textit{Yaw Jump} use a time phase variable $\Phi$ \cite{10.1145/3197517.3201311} to motivate the position of feet contacts to produce jump. 
The summary of observations are reported in Table\ref{tab:observations}

\begin{table}[h]
\centering
\caption{Observation Space for RL Policy}
\begin{tabular}{ll}
\toprule
\textbf{Component} & \textbf{Description} \\
\midrule
\multicolumn{2}{c}{\textit{Common Observations (used in all tasks)}} \\
\midrule
$\boldsymbol{\omega}^\text{base}_t$ & Base angular velocity \\
$\mathbf{g}_t$ & Gravity vector projected in base frame \\
$\mathbf{q}_t$ & Joint positions \\
$\dot{\mathbf{q}}_t$ & Joint velocities \\
$\mathbf{a}_t$ & Last applied actions \\
\midrule
\multicolumn{2}{c}{\textit{Task-Specific Observations}} \\
\midrule
\textbf{Forward \& Yaw Jump} & $\Phi$ (phase), $\mathbf{a}_{t-1}$ (last-last action) \\
\textbf{Velocity Tracking} & $v^{x}_\text{cmd}$, $v^{y}_\text{cmd}$, $\omega^z_\text{cmd}$ \\
\textbf{Attitude Tracking} & $\phi_\text{cmd}$(roll), $\psi_\text{cmd}$(pitch) \\
\textbf{Humanoid Velocity Tracking} & $\Phi$ (phase), $q^{\text{ref}_{\text{upper}}}$(upperbody dof reference)\\
\bottomrule
\end{tabular}
\label{tab:observations}
\end{table}

\subsubsection{Rewards}

We summarize the weights of all reward terms used in tasks during policy training and evaluation. Quadruped task rewards are detailed in Table~\ref{tab:reward_quadruped}, covering different locomotion and agility objectives including block jumping, yaw jumping, tracking, and agile movement. Humanoid task rewards are shown in Table~\ref{tab:reward_humanoid}, with emphasis on gait stabilization, symmetry, and whole-body coordination.

\begin{table}[h]
\centering
\small
\caption{Reward terms for quadruped tasks}
\label{tab:reward_quadruped}
\renewcommand{\arraystretch}{1.2}
\begin{tabular}{lcccc}
\toprule
\textbf{Term}                  & \textbf{Block Jump} & \textbf{Yaw Jump} & \textbf{Roll/Pitch Track} & \textbf{Agile Loco} \\
\midrule
\multicolumn{5}{l}{\textbf{Task Reward}} \\
\midrule
Body position                  &   2.0    &   2.0    &   –      &   –      \\
Body orientation               &    –     &   2.0    &   3.0    &   –      \\
Body linear velocity           &    –     &    –     &   –      &   2.0    \\
Body angular velocity          &    –     &    –     &   –      &   1.0    \\
Feet height                    &   3.0    &   3.0    &   –      &   –      \\

\midrule
\multicolumn{5}{l}{\textbf{Penalties \& Regularization}} \\
\midrule
Action rate             & –1e-3    & –1e-3    & –1e-3    & –1e-2    \\
Slippage                       & –3.0     & –3.0     & –1e-1    & –      \\
In-air contact                    & –3.0     & –3.0     &   –      &   –      \\
% Landing contact                &   –      &   –      &   –      & –5e-2    \\
Foot spacing                   & –2.5     & –2.5     & –0.5     &   –      \\
Non-foot contact               &   –      & –0.3     &   –      & –1e-1    \\
Torque penalty                 &   –      &   –      & –2e-4    & –2e-4    \\
Acceleration penalty           &   –      &   –      & –2.5e-7  & –2.5e-7  \\
Velocity penalty               &   –      &   –      & –1e-4    & –1e-4    \\
Symmetry bonus                 &  2.0     &  2.0     &   –      &   –      \\
Base-height reference          &   –      &   –      &  1.0     &  2.0     \\
% Collision penalty              &   –      &   –      & –1e-1    &   –      \\
Joint-limit violation          &   –      &   –      & –10.0    & –10.0    \\
\bottomrule
\end{tabular}
\end{table}

\begin{table}[h]
\centering
\small
\caption{Reward terms for humanoid velocity tracking}
\label{tab:reward_humanoid}
\renewcommand{\arraystretch}{1.2}
\begin{tabular}{l c l c}
\toprule
\textbf{Term} & \textbf{Weight} & \textbf{Term} & \textbf{Weight} \\
\midrule
\multicolumn{4}{l}{\textbf{Task Reward}} \\
\midrule
Linear-velocity tracking  & 1.0   & Angular-velocity tracking & 1.0   \\
Waist-joint tracking      & 0.5     \\
\midrule
\multicolumn{4}{l}{\textbf{Penalties \& Regularization}} \\
\midrule
Action-rate               & –0.1  & Vertical-vel              & –2.0  \\
Lateral-ang-vel           & –0.05 & Orientation               & –1.5  \\
Torque                    & –1e-5 & Acceleration              & –2.5e-7 \\
Velocity                  & –1e-3 & Contact-no-vel            & –0.2  \\
Feet-orientation          & –2.0  & Close-feet                & –10.0 \\
Joint-limit violation     & –5.0  & Base-height reference     & –10.0 \\
Contact                   & –0.20 & Feet-heading alignment    & –0.25 \\
Hip-position              & –1.0  & Stance-tap                & –5.0  \\
Stance-root               & –5.0  & Stance-symmetry           & –0.5  \\
Survival (“alive”)        & 0.15  & Contact bonus             & 0.18  \\
\bottomrule
\end{tabular}
\end{table}

\subsubsection{Domain Randomization}
We use two Domain Randomization ranges. The nominal range, that is used by the vanilla baseline, \method and \methodactive. However it should be noted that the motor parameter ranges $\kappa_{\text{Hip}},\kappa_{\text{Thigh}},\kappa_{\text{Calf}}$ corresponding to the Hip, Thigh and Calf are not used for the vanilla baseline. Second, we have the Heavy Range, where the ranges are almost double compared to the nominal range and is used by the \textit{Heavy DR} baseline. The exact ranges are summarized in Table \ref{tab:DR}

\begin{table}[h]
    
    \centering
    \footnotesize
    \setlength{\tabcolsep}{2pt} % Reduce column spacing
    \caption{Domain Randomization Ranges}
    \resizebox{0.75\textwidth}{!}{
    \begin{tabular}{ccc}
        \hline
        Term & Nominal Range & Heavy Range \\ 
        \hline
        \multicolumn{3}{c}{Dynamics Randomization} \\
        \hline
        Base CoM offset($\mathrm{m}$) & $\mathcal{U}(-0.1,0.1) $ & $\mathcal{U}(-0.2,0.2) $ \\
        Base mass($\times$ default)Kgs  & $\mathcal{U}(0.8,1.2)$  & $\mathcal{U}(0.6,1.4) $ \\
        Base Inertia offset($Kg\mathrm{m}^2$) & $\mathcal{U}(-0.05,0.05)$ & $\mathcal{U}(-0.05,0.3) $ \\
        $\kappa_{\text{Hip}}$ & $\mathcal{U}(22,24)$ & - \\
        $\kappa_{\text{Thigh}}$ & $\mathcal{U}(24,26)$ & - \\
        $\kappa_{\text{Calf}}$ & $\mathcal{U}(22,24)$ & - \\
        % Control delay & $\mathcal{U}(20,40) \mathrm{ms}$ \\
        \hline
         Term & Value\\ 
        \hline
        \multicolumn{2}{c}{Common Ranges} \\
        \hline
        P Gain & $\mathcal{U}(0.9,1.1)$\\
        D Gain & $\mathcal{U}(0.9,1.1)$\\
        Torque RFI  & $0.1 \times$ torque limit $\mathrm{N} \cdot \mathrm{m}$ \\
        \hline
    \end{tabular}}
    \label{tab:DR}
\end{table}
\vspace{-2mm}
\subsection{Parameter Identification Result Analysis} \label{param_analysis}
We evaluate the identified physical and actuator parameters of the Go2 robot with a $4.7$kg payload mounted at the lower rear side of the base. The Table~\ref{tab:sysid_comparison} compares the identified values against the default parameters without payload. The identified model captures key physical changes introduced by the $4.7,\mathrm{kg}$ rear-mounted payload. The estimated mass increases by approximately $2.44,\mathrm{kg}$, partially compensating for the added load. The center of mass shifts rearward and downward, consistent with the payload’s mounting location, and is essential for accurate contact force modeling and stability. Among the inertial parameters, we observe a reasonable increase in $\textbf{I}_{zz}$, likely resulting from the added mass distribution around the yaw axis. However, the increases in $\textbf{I}_{xx}$ and $\textbf{I}_{yy}$ are unexpectedly large and not fully supported by the payload geometry. This suggests possible overfitting or parameter coupling due to insufficient excitation in the pitch and roll directions—an inherent challenge for quadruped systems. Nonetheless, the identified parameters yield improved trajectory prediction and real-world policy transfer performance, indicating that the model captures useful aspects of the true dynamics. The actuator tanh gains reveal strong saturation effects at high-torque range. With gains around 25 for the thigh and calf joints, torque output saturates more gradually, resulting in a $20-26$\% reduction near the maximum torque limits. Modeling this nonlinearity is critical for improving sim-to-real fidelity in high-torque tasks such as dynamic locomotion and jumping with payloads.

\begin{table}[h]
\centering
\caption{Comparison of Default and Identified Parameters of Go2}
\label{tab:sysid_comparison}
\small
\begin{tabular}{lccccccccccc}
\toprule
\textbf{Setting} & \textbf{Mass} & \textbf{CoM$_x$} & \textbf{CoM$_y$} & \textbf{CoM$_z$} & $\textbf{I}_{xx}$ & $\textbf{I}_{yy}$ & $\textbf{I}_{zz}$ & $\kappa_{\text{Hip}}$ & $\kappa_{\text{Thigh}}$ & $\kappa_{\text{Calf}}$ \\
\midrule
Default & 6.921 & 0.021 & 0.000 & -0.005 & 0.025 & 0.098 & 0.107 & — & — & — \\
Payload & 9.363 & 0.004 & -0.005 & -0.020 & 0.391 & 0.515 & 0.396 & 22.553 & 24.969 & 23.523 \\
\bottomrule
\end{tabular}
\end{table}

% \subsection{Attitude Tracking Results}\label{attitdue_track}

% Figure~\ref{fig:attitude_tracking} shows pitch tracking performance in an attitude control task, emphasizing the sim-to-real consistency of SPI-Active policies. The SPI-Active policy (left) exhibits a trajectory that closely matches the simulated response, whereas the vanilla policy (right) shows larger deviations from simulation. This demonstrates that SPI-Active achieves a smaller sim-to-real gap.

% \begin{figure*}[h]
%     \centering
%     \begin{subfigure}[t]{0.5\textwidth}
%         \centering
%         \includegraphics[height=1.4in]{figs/spi-active.pdf}
%         \caption{SPI-Active}
%     \end{subfigure}%
%     ~ 
%     \begin{subfigure}[t]{0.5\textwidth}
%         \centering
%         \includegraphics[height=1.4in]{figs/vanilla.pdf}
%         \caption{Vanilla}
%     \end{subfigure}
%     \caption{Open-loop attitude tracking results. Comparison between SPI-Active and vanilla policies in both simulation and real-world execution. SPI-Active yields a closer match to simulation, suggesting improved sim-to-real consistency.}
%     \label{fig:attitude_tracking}
% \end{figure*}

\end{document}